\documentclass{article} %
\usepackage{colm2024_conference}

\usepackage{booktabs}
\usepackage{graphicx}
\usepackage{enumitem}
\usepackage{wrapfig}
\usepackage{algorithm}
\usepackage{algpseudocode}
\usepackage{wrapfig}
\usepackage{float}
\usepackage{microtype}
\usepackage{amsmath}
\usepackage{amssymb}
\usepackage{colortbl}
\usepackage[utf8]{inputenc}
\usepackage{caption}
\usepackage{subcaption}
\usepackage{xcolor}
\usepackage{setspace}
\usepackage{url}
\usepackage{multirow}
\usepackage{colortbl}
\usepackage{tabularx}
\usepackage{blindtext}
\usepackage{pgfplots}
\pgfplotsset{compat=1.18}
\usepackage{tikz}
\usetikzlibrary{er, positioning, bayesnet}
\usepackage{makecell}
\usepackage{tipa}
\usepackage{siunitx}
\usepackage{nicefrac}
\usepackage{tocloft}
\usepackage{listings}
\usepackage[raster, skins]{tcolorbox} %
\usepackage{xltabular}
\usepackage{adjustbox}
\usepackage{xurl}
\usepackage{longtable}  

\usepackage{amsmath,amsfonts,bm}

\def\eqref#1{equation~\ref{#1}}

\def\1{\bm{1}}

\DeclareMathAlphabet{\mathsfit}{\encodingdefault}{\sfdefault}{m}{sl}
\SetMathAlphabet{\mathsfit}{bold}{\encodingdefault}{\sfdefault}{bx}{n}

\definecolor{lightgray}{rgb}{0.9,0.9,0.9}

\usepackage{afterpage}
\usepackage{cuted}
\usepackage{stfloats}
\usepackage{ulem}           
\usepackage{soul}           
\usepackage{tablefootnote}  
\usepackage{comment}        

\usepackage{pifont}         
\usepackage{wasysym}

\usepackage{longtable}
\usepackage{array}
\usepackage{caption}  
\usepackage{ragged2e} 

\usepackage{listings}
\usepackage{xcolor}
\usepackage{graphicx}
\usepackage{wrapfig}
\usepackage{amsthm}

\newtheorem{theorem}{Theorem}[section]

\definecolor{lightblue}{RGB}{173,216,230}
\definecolor{lightgreen}{RGB}{144,238,144}
\definecolor{lightpink}{RGB}{255, 228, 225}
\definecolor{lightred}{RGB}{255,182,193}
\definecolor{lightyellow}{RGB}{255,255,224}
\definecolor{lightpurple}{RGB}{221,160,221}
\definecolor{lightgray}{RGB}{211,211,211}
\definecolor{lightorange}{RGB}{255,218,185}
\definecolor{lightpeach}{rgb}{1.0, 0.882, 0.788}
\definecolor{lightcyan}{rgb}{0.8196, 0.9725, 0.9804}
\definecolor{sh_blue}{rgb}{0,0.60,0.93}
\definecolor{sh_red}{rgb}{0.8627, 0.3098, 0.3176}
\definecolor{highlight}{RGB}{255,255,0}
\definecolor{warning}{RGB}{255,99,71}
\definecolor{success}{RGB}{50,205,50}
\definecolor{info}{RGB}{30,144,255}
\definecolor{top1}{RGB}{255,179,179}
\definecolor{top2}{RGB}{255,217,179}
\definecolor{top3}{RGB}{255,255,179}
\definecolor{textblue}{RGB}{94,159,220} 
\definecolor{textgreen}{RGB}{59,125,35} 
\definecolor{textorange}{RGB}{192,80,21} 
\definecolor{tagred}{RGB}{196,15,15} 
\definecolor{tagblue}{RGB}{33,95,154} 
\definecolor{teaserblue}{RGB}{33,95,154} 
\definecolor{teasergree}{RGB}{57,158,163} 
\definecolor{teaserpurpe}{RGB}{105,111,173} 

\definecolor{primary}{RGB}{70,130,180}
\definecolor{secondary}{RGB}{119,136,153}
\definecolor{accent}{RGB}{255,140,0}

\definecolor{customblue}{HTML}{E7EFFA}
\definecolor{custompink}{HTML}{F7E1ED}

\renewcommand{\arraystretch}{1.25}

\makeatletter
\DeclareRobustCommand\onedot{\futurelet\@let@token\@onedot}
\def\@onedot{\ifx\@let@token.\else.\null\fi\xspace}

\makeatother

\title{
VisionCreator-R1: A Reflection-Enhanced Native Visual-Generation Agentic Model}

\author{%
Jinxiang Lai$^{2*}$, Wenzhe Zhao$^{1*}$, Zexin Lu$^{1*}$, Hualei Zhang$^{2}$, Qinyu Yang$^{1}$, Rongwei Quan$^{1}$, Zhimin Li$^{1}$, Shuai Shao$^{1}$,  Song Guo$^{2\S}$, Qinglin Lu$^{1\dagger}$\\
{\small{$^1$Tencent Hunyuan,
$^2$Hong Kong University of Science and Technology}}\\
{\small{$^*${Equal contribution}, $^\S${Corresponding Author}, $^\dagger${Project lead}}}
}

\begin{document}

\maketitle

\begin{abstract}
Visual content generation has advanced from single-image to multi-image workflows, yet existing agents remain largely plan-driven and lack systematic reflection mechanisms to correct mid-trajectory visual errors.
To address this limitation, we propose \textbf{VisionCreator-R1}, a native visual generation agent with explicit reflection, together with a \textbf{Reflection--Plan Co-Optimization (RPCO)} training methodology.
Through extensive experiments and trajectory-level analysis, we uncover reflection--plan optimization asymmetry in reinforcement learning (RL): planning can be reliably optimized via plan rewards, while reflection learning is hindered by noisy credit assignment.
Guided by this insight, our RPCO first trains on the self-constructed \textbf{VCR-SFT} dataset with reflection-strong single-image trajectories and planning-strong multi-image trajectories, then co-optimization on \textbf{VCR-RL} dataset via RL.
This yields our unified VisionCreator-R1 agent, which consistently outperforms Gemini2.5Pro on existing benchmarks and our \textbf{VCR-bench} covering single-image and multi-image tasks.
\end{abstract}

\section{Introduction}
AI-assisted visual generation has rapidly evolved from single-image generation \citep{ho2020denoising,ramesh2022hierarchical,flux2024,lin2024open} to complex, multi-image and video workflows \citep{wu2025automated,xiao2025captain,xu2025mm,xue2025comfybench,guo2025comfymind}. 
Existing methods for visual generation have made some progress, yet they still face numerous challenges. Workflow-specific agents \citep{wu2025automated,xiao2025captain,xu2025mm} rely on hand-crafted pipelines tailored to narrow domains and struggle to adapt when task structure or tools change. Workflow-guided agents \citep{xue2025comfybench,guo2025comfymind} orchestrate external tools through carefully designed system prompts and coordination logic, but this hard-coded control flow prevents end-to-end optimization of understanding, planning, and generation.
This inevitably hinders the seamless alignment of complex user instructions with precise visual outputs and precludes the system from learning globally effective policies. 

\begin{figure*}[t]
\vspace{-2mm}
\centering
\includegraphics[width=0.99\textwidth]{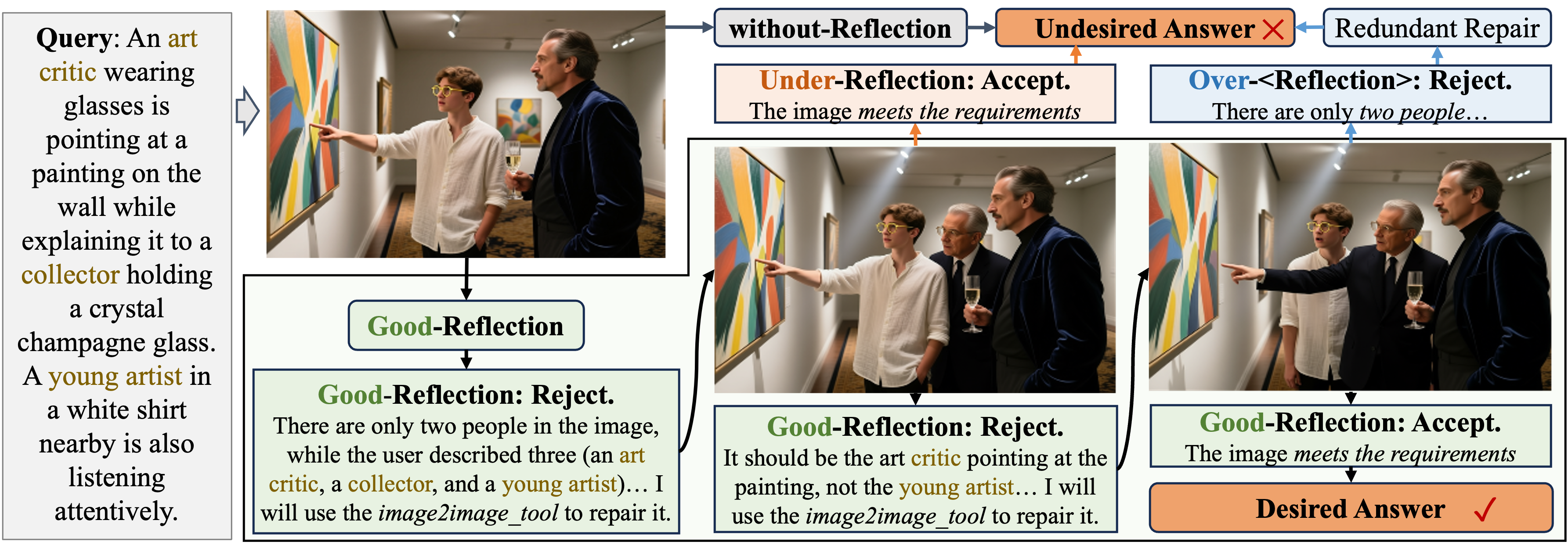}
\vspace{-2mm}
\caption{Comparison between without-Reflection, with {Good-Reflection}, Under-Reflection and Over-Reflection.}
\vspace{-3mm}
\label{fig:motivation}
\end{figure*}

Motivated by the need for a unified and differentiable architecture,
a recent native visual generation agent named VisionCreator \citep{lai2026visioncreator} integrates Understanding, Thinking, Planning, and Creation (UTPC) into a single trainable framework, outperforming ``VLM~+~tools'' systems on multi-step visual generation tasks.
Yet these agents remain fundamentally \textit{plan-driven}, where designed system prompts or training signals prioritize the rationality of plans and tool calls. This exclusive focus on procedural correctness precludes a retrospective mechanism for \textit{structured reflection and self-correction}. Consequently, minor deviations in early stages propagate unchecked, leading to severe error accumulation in long-horizon, multi-image workflows.
As shown in Fig.~\ref{fig:motivation} (a), \textbf{without-Reflection} outputs an Undesired Answer, which will cause error accumulation in subsequent trajectories.

Recent works such as JarvisEvo \cite{lin2025jarvisevo} and ReasonEdit \cite{yin2025reasonedit} demonstrate that introducing reflection into \textit{single-image editing} can improve visual quality. However, these settings involve short-horizon tasks with minimal planning requirements. In contrast, native visual agents must handle diverse workloads spanning single-image and multi-image, all of which rely heavily on robust planning. This raises a new challenge: how to \textit{co-optimize planning and reflection} within a native visual generation agent to improve long-horizon behavior.

In this paper, we systematically study reflection as a trainable behavior inside a native visual generation agent.
We seek to answer three core questions:
(i) \textit{Why is reflection necessary for visual generation agents?}
(ii) \textit{How can effective reflection be rewarded without encouraging superficial or ungrounded edits?}
(iii) \textit{How can we effectively jointly optimize planning and reflection within a unified framework?}
To this end, we introduce \textbf{VisionCreator-R1} and propose a \textbf{Reflection--Plan Co-Optimization (RPCO)} methodology that progresses from decoupled to fused training.

\noindent \textbf{Stage~1: Isolating reflection on single-image tasks.}
To ensure that performance gains can be unambiguously attributed to reflection, we begin with single-image generation tasks where the demand for explicit planning is minimal.
Under this setting, the resulting \textit{Strong-Reflection} model surpasses Gemini2.5Pro on single-image benchmarks, establishing that reflection is both learnable and beneficial for visual generation.

\noindent \textbf{Stage~2: Extending to multi-image tasks and revealing optimization asymmetry.}
Motivated by the need to handle realistic, multi-image tasks, we then extend reflection-optimized models to multi-image workflows via Reinforcement Learning (RL).
While overall task performance and plan scores improve, reflection quality shows no improvement in multi-image trajectories.
Through analysis, we identify a fundamental \textit{optimization asymmetry} between planning and reflection:
plan rewards directly supervise planning decisions, whereas reflection rewards are computed solely from post-reflection visual outcomes and are influenced by planning quality, tool execution fidelity, and image generation stochasticity.
This noisy credit assignment makes reflection difficult to optimize under long-horizon RL.

\noindent \textbf{Stage~3: RPCO methodology.}
To address this challenge, we propose an advantage-complementary training strategy.
As depicted in Fig.~\ref{fig:framework} (b), we perform supervised fine-tuning (SFT) on the self-constructed \textbf{VCR-SFT dataset} with a mixture of reflection-strong single-image trajectories and planning-strong multi-image trajectories, yielding a balanced Reflection--Plan SFT model.
Starting from this initialization, we then apply multi-task RL on our \textbf{VCR-RL dataset}.
In this regime, planning continues to improve under reliable plan-level rewards, while reflection capability learned during SFT is preserved.
The final \textit{VisionCreator-R1} model achieves consistent gains over Gemini2.5Pro on existing benchmarks and our \textbf{VCR-bench} covering single-image and multi-image tasks.

In summary, our contributions are as follows:
(i) We identify a critical transferability gap where reflection capabilities learned in single-image settings fail to generalize to multi-image scenarios. Through theoretical analysis, we attribute this failure to a fundamental structural variance asymmetry: while planning benefits from stable rewards, reflection optimization in long-horizon tasks is hindered by collapsed signal-to-noise ratios due to trajectory-level stochasticity.
(ii) We propose VisionCreator-R1, a native visual generation agent trained via a novel RPCO paradigm. Guided by our theoretical analysis, RPCO adopts a "decoupled-then-fused" strategy, isolating reflection learning in low-noise settings before synergizing it with planning via multi-task RL.
(iii) We construct VCR-SFT and VCR-RL datasets to support RPCO methodology.
And we propose VCR-Bench, a VLM-graded benchmark covering single-image, multi-image, and image-to-image tasks with multiple evaluation points per query, enabling standardized assessment.

\section{Related Works}
\label{sec:related}

\subsection{Image Generation}
Modern image generation models, such as autoregressive models~\citep{chen2020generative,fan2024fluid,han2024infinity,tian2024visual,sun2024autoregressive,pang2024next} and diffusion models~\citep{ho2020denoising,ramesh2022hierarchical,flux2024,saharia2022photorealistic,lin2024open}, offer single-step text-to-image and image-to-image capabilities. However, they do not themselves provide mechanisms for task decomposition, tool orchestration, or structured reflection over intermediate results. VisionCreator-R1 treats these generators as callable tools, focusing on how to \textit{plan, reflect, and correct} across long visual trajectories.

\subsection{Visual Generation Agents}
\textbf{Workflow-Specific Agents} \citep{wu2025automated,xiao2025captain,xu2025mm} design domain-specific pipelines for tasks like movie-style storyboarding or episodic content creation. They struggle when task structure or tools deviate from the handcrafted templates.
\textbf{Workflow-Guided Agents} \citep{xue2025comfybench,guo2025comfymind,xu2025comfyui,huang2025comfygpt} focus on generating ComfyUI-compatible node graphs. They advance automated workflow construction, but are tied to a specific execution backend, and typically realise control flow via static prompts and rules.
\textbf{Native Visual Agents} such as VisionCreator~\citep{lai2026visioncreator} unify UTPC in a trainable framework, outperforming ''VLM+tools'' baselines. However, existing UTPC agents are still \textit{plan-driven}: training and evaluation primarily supervise whether plans are reasonable, and tools are called valid. They lack an explicit, structured \textit{reflection} path to evaluate intermediate images and trigger self-corrections. Our VisionCreator-R1 builds on this line of work but introduces reflection and studies how to co-optimise reflection and planning.

\section{Empirical and Analytical Foundation}
\label{sec:motivation}
\subsection{Agentic Framework}
\label{sec:framework}
As illustrated in Fig.~\ref{fig:framework} (a), we instantiate the UTPCR framework to construct a dynamic loop of Act–Reflect–Think–Act. Unlike traditional feed-forward pipelines, our agent incorporates an explicit \textit{reflection path} to enable self-correction. In this loop, the agent utilizes its visual understanding capabilities to: (i) scrutinize intermediate visual results for quality and consistency, (ii) detect deviations from the user's instructions and prior context, and (iii) formulate corrective plans, triggering targeted edits or re-generation, to ensure the trajectory aligns precisely with the desired outcome.

\begin{wrapfigure}{r}{0.37\textwidth}
\vspace{-3mm}
\centering
\includegraphics[width=0.9\linewidth]{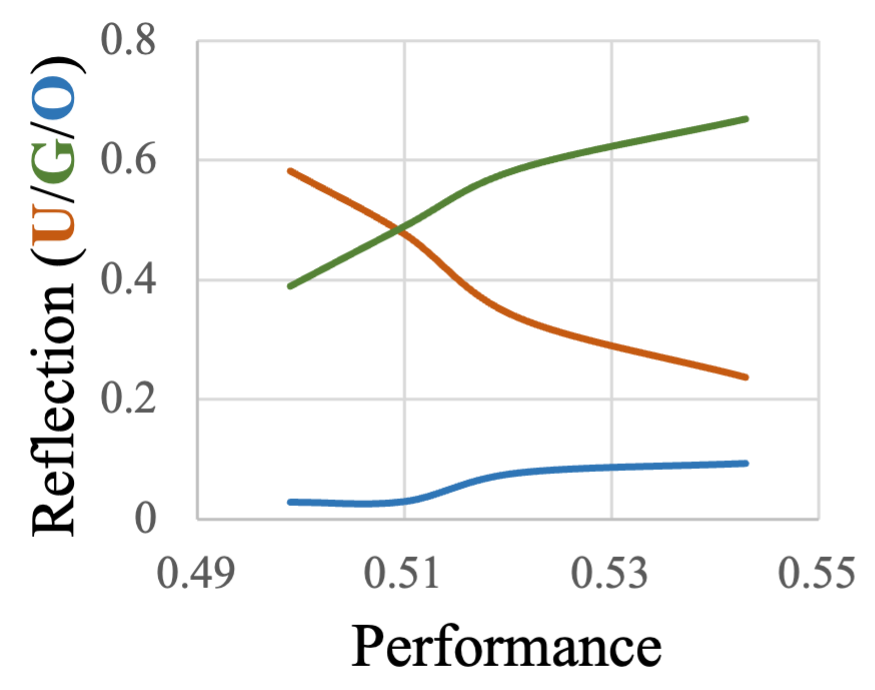}
\vspace{-2mm}
\caption{Performance - Reflection Quality.}
\vspace{-2mm}
\label{fig:exp_single}
\end{wrapfigure}

\subsection{Reflection Optimization on Single-Image Tasks}
\label{sec:reflection}
To cleanly study reflection independent of complex planning, we begin with single-image tasks involving image generation and editing tools. 
First, we construct UTPCR-format reflection trajectories by querying Gemini2.5Pro.
Second, we perform SFT on this corpus to endow the agent with explicit reflection structures and basic reflection behaviors in the UTPCR trajectory.
Finally, we introduce RL driven by the visual reflection reward $R_{\text{reflect}}$ defined in Eq.~\ref{reflect_reward} to further optimize the effectiveness of reflection in improving final visual quality.
Through this two-stage SFT+RL training, we obtain a \textit{Strong-Reflection model} that consistently outperforms Gemini2.5Pro on single-image tasks, providing clear evidence that our reflection mechanism and reward design are effective for visual generation agents.

\paragraph{Single-Image Reflection Experiments.}
Fig.~\ref{fig:exp_single} shows that increasing the \textbf{Good-Reflection} percentage leads to the improvement of final visual generation.
And Tab.~\ref{tab:single_image} summarizes the detailed experiments on the single-image task in VCR-Bench. 
Several observations can be made from it:
(i) Merely adding a reflection structure is insufficient to guarantee performance gains. Gemini2.5Pro and Qwen3VL32B show only marginal improvements over the baseline. 
(ii) Reflection learned purely via SFT (SFT-Reflection model) leads to degraded performance and a strong bias toward under-reflection, indicating that the model becomes overly conservative.
(iii) After introducing RL with the proposed $R_{\text{reflect}}$, both the single-img performance and the proportion of good-reflection increase steadily from Reflection-s30 to Strong-Reflection (i.e., Reflection-s180). Meanwhile, under-reflection decreases substantially, and over-reflection remains at a moderate level.
This quantitatively validates that for single-image tasks, the visual reflection reward successfully incentivizes grounded reflection, directly leading to tangible gains in visual fidelity.

\begin{wraptable}{r}{0.55\columnwidth}
\centering
\renewcommand{\tabcolsep}{1.1pt}
\renewcommand{\arraystretch}{1.1}
\small
\caption{Reflection experiment on Single-Img task in VCR-Bench. Ref. Quality (U/G/O): reflection quality with under-/good-/over-reflection percentages. And all models' plan score on the single-image task is larger than 99.5\%.}
\label{tab:single_image}
\begin{tabular}{l|c|c}
\hline
\textbf{Model} & \textbf{Single-Img} & \textbf{Ref. Quality (U/G/O)} \\
\hline
Qwen-Image-Fast & 0.505 & -- \\
\hline
Gemini2.5Pro & 0.515 & 32.0\%/55.4\%/12.6\% \\
Qwen3VL32B & 0.500 & 41.3\%/49.7\%/9.0\% \\
\hline
SFT-Reflection & 0.499 & 58.2\%/38.9\%/2.9\% \\
Reflection-s30 & 0.510 & 47.7\%/48.9\%/3.3\% \\
Reflection-s130 & 0.521 & 33.7\%/58.5\%/7.8\% \\
Strong-Reflection & \textbf{0.543} & {23.7\%}/\textbf{66.9\%}/9.4\% \\
\hline
\end{tabular}
\vspace{-1mm}
\end{wraptable}

\subsection{Reflection and Planning in Multi-Image Tasks}
\label{sec:conflict}

\paragraph{Direct transfer from single-image reflection to multi-image via RL.}
We first examine whether a reflection-optimized model trained on single-image tasks can be directly extended to multi-image settings.
Starting from the \textit{Strong-Reflection} model---which achieves the highest single-image performance, we further perform RL training on the multi-image dataset, obtaining the \textit{Reflection-Plan Conflict} model.
Multi-image tasks require the agent to reason over longer horizons, construct reference images, maintain cross-image consistency, and follow structured narratives.
However, as shown in Tab.~\ref{tab:multi_image}, direct RL transfer yields good improvements in multi-image performance (0.617), but still underperforms strong planners such as Gemini2.5Pro (0.649).

\begin{wraptable}{r}{0.55\columnwidth}
\centering
\vspace{-3mm}
\renewcommand{\tabcolsep}{1.8pt}
\renewcommand{\arraystretch}{1.1}
\small
\caption{Transfer from single-image reflection to multi-image tasks. Model: SR is Strong-Reflection model, RPC is Reflection-Plan Conflict model, GPro is Gemini2.5Pro.}
\label{tab:multi_image}
\vspace{-1mm}
\begin{tabular}{l|c|c|c}
\hline
\textbf{Model} & \textbf{Multi-Img} & \textbf{Ref. Quality (U/G/O)} & \textbf{Plan Score} \\
\hline
SR  & 0.565 & 68.6\%/21.6\%/9.8\% & 0.8285 \\
RPC & 0.617 & 73.9\%/16.5\%/9.7\% & 0.8431 \\
GPro & 0.649 & 61.6\%/31.8\%/6.6\% & 0.9667 \\
\hline
\end{tabular}
\vspace{-1mm}
\end{wraptable}

\paragraph{Trajectory-level analysis of reflection and planning.}
Compared with the Strong-Reflection model, the Reflection-Plan Conflict model exhibits:
(i) only marginal improvement in plan score (0.8431 vs.\ 0.8285), and
(ii) a further degradation in reflection quality, with the proportion of high-quality reflections decreasing from 21.6\% to 16.5\%.
This indicates that single-image reflection capabilities cannot be directly transferred to multi-image workflows via RL. While this approach improves planning performance, it leads to a degradation in reflection quality.

\begin{figure*}[ht]
\vspace{-2mm}
\centering
\includegraphics[width=0.99\textwidth]{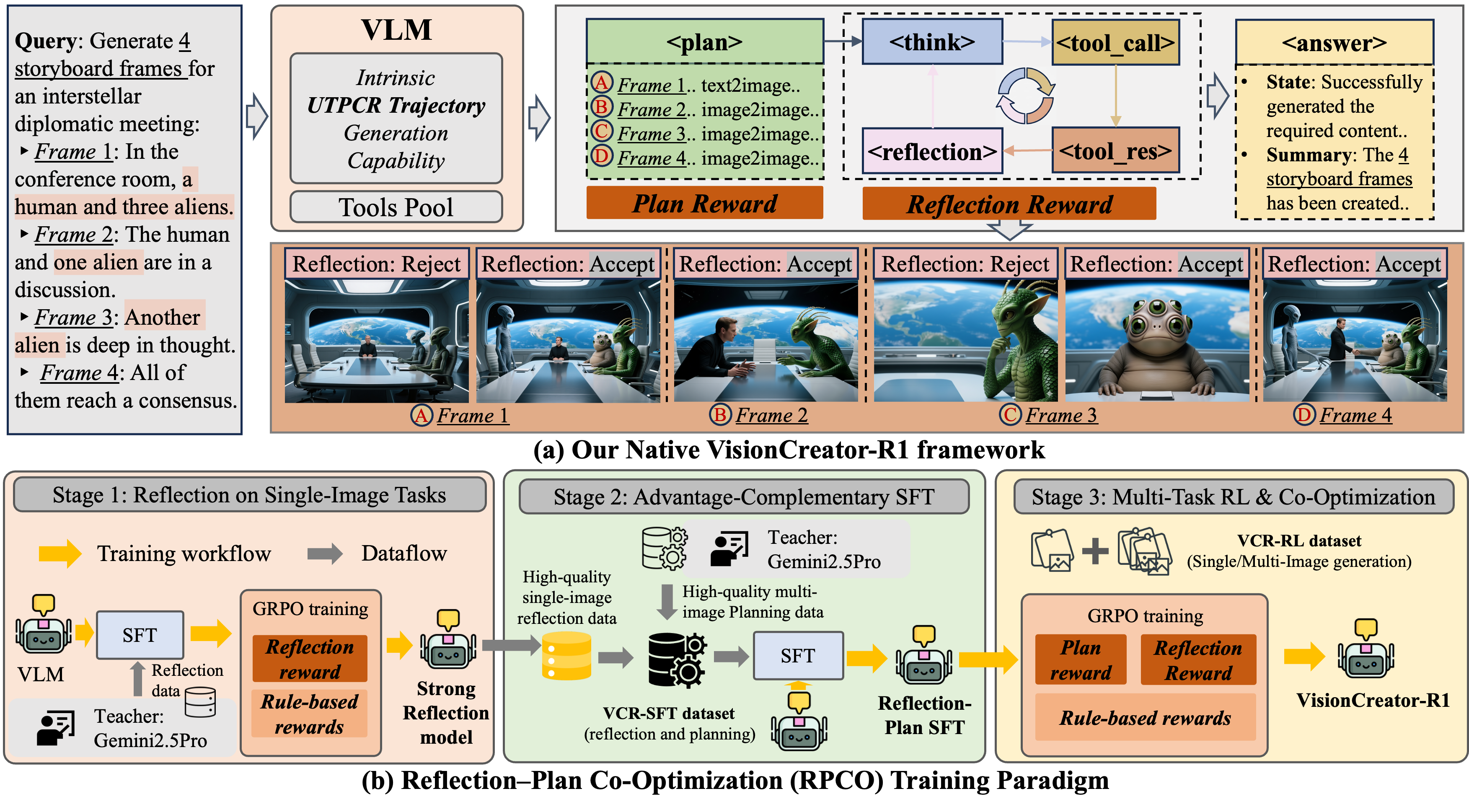}
\vspace{-3mm}
\caption{(a) Our Native VisionCreator-R1 framework. (b) Reflection–Plan Co-Optimization (RPCO) Training Paradigm.}
\vspace{-2mm}
\label{fig:framework}
\end{figure*}

\paragraph{Why reflection capabilities are difficult to optimize in multi-image tasks.}
Our Theorem~\ref{thm:variance_asymmetry} reveals that the difficulty arises from a fundamental \textit{structural variance asymmetry} between planning and reflection rewards in the GRPO optimization objective.

\begin{theorem}[Structural Variance Asymmetry in Multi-Image GRPO]
\label{thm:variance_asymmetry}
Consider the GRPO optimization objective and its stochastic gradient estimator for a single token generation step $t$ in trajectory $i$.
Given state $s = (q, o_{i,<t})$ and sampled action $a = o_{i,t}$, the gradient estimator is
\begin{equation}
\hat{g} =
\left[
\hat{A}_{i,t}
+ \beta \left( \frac{\pi_{\mathrm{ref}}(a \mid s)}{\pi_\theta(a \mid s)} - 1 \right)
\right]
\nabla_\theta \log \pi_\theta(a \mid s),
\end{equation}
where $\hat{A}_{i,t}$ is normalized advantage derived from trajectory-level reward,
$\pi_\theta$ is the current policy, $\pi_{\mathrm{ref}}$ is a fixed reference policy, and $\beta$ is the KL regularization coefficient.

The total variance of $\hat{g}$ admits the following decomposition:
\begin{equation}
\Sigma := \text{Var}(\hat{g})
=
\underbrace{\mathbb{E}_{s,a} \left[ \text{Var}_{\tau \mid s,a}(\hat{g}) \right]}_{\Sigma_\tau}
+
\underbrace{\mathbb{E}_s \left[ \text{Var}_{a \mid s}
\left( \mathbb{E}_{\tau \mid s,a}[\hat{g}] \right) \right]}_{\Sigma_a},
\end{equation}
where $\tau$ denotes the stochastic trajectory.
$\Sigma_\tau$ captures the variance induced by environmental dynamics and reward stochasticity,
while $\Sigma_a$ captures the variance induced by stochastic action sampling.

Then, a fundamental \textbf{structural variance asymmetry} holds between planning and reflection objectives:
\begin{itemize}
    \item For \textbf{planning rewards} $R_{\mathrm{plan}}$, which are evaluated deterministically conditioned on the action,
    the conditional trajectory variance is negligible,
    \begin{equation}
        \text{Var}_{\tau \mid s,a}(\hat{A}_{i,t}) \approx 0
        \quad \Rightarrow \quad
        \Sigma_\tau \approx 0,
    \end{equation}
    and the gradient variance is dominated by $\Sigma_a$, yielding a stable optimization regime.
    
    \item For \textbf{reflection rewards} $R_{\mathrm{reflect}}$ in multi-image generation,
    which depend on stochastic downstream visual outcomes,
    the conditional trajectory variance is strictly positive and often dominant,
    \begin{equation}
        \text{Var}_{\tau \mid s,a}(\hat{A}_{i,t}) \gg 0
        \quad \Rightarrow \quad
        \Sigma_\tau \gg \Sigma_a.
    \end{equation}
\end{itemize}

Consequently, reflection optimization in multi-image GRPO suffers from a collapsed Signal-to-Noise Ratio,
making it fundamentally harder to optimize than planning.
\end{theorem}

\noindent {Intuitively, for planning}, the reward signal $R_{\text{plan}}$ evaluates the action $a_{\text{plan}}$ directly via a deterministic LLM-based evaluator (assessing logical coherence and tool matching). Since the planning rewards $R_{\mathrm{plan}}$ excludes the stochastic image generation process, environmental noise is negligible ($\Sigma_\tau \approx 0$). Consequently, the gradient variance is dominated by $\Sigma_a$, which provides a stable exploration signal for the optimizer.

\noindent {In contrast}, the optimization of {reflection} faces severe challenges due to the specific nature of multi-image tasks. First, unlike single-image generation, multi-image tasks require maintaining long-horizon properties like cross-image consistency and narrative structure, leading to an exponential expansion of the decision space. Second, the reflection reward $R_{\text{reflect}}$ relies on the \textit{post-reflection visual outcome}, meaning the conditional trajectory $\tau|a_{\text{reflect}}$ encompasses the entire generation chain. This couples the reflection action with the high stochasticity of image generation process and tool execution, resulting in a scenario where trajectory variance dominates ($\Sigma_\tau \gg \Sigma_a$).
As a result, the variance in the reflection rewards is driven primarily by generation noise rather than the quality of the reflection action itself ($\Sigma_\tau \gg \Sigma_a$). This noise drowns out the useful signal, making it intractable for the agent to distinguish between erroneous reflection and the inherent noise of image generation. 

Consequently, reflection capabilities become difficult to optimize in multi-image tasks, rendering direct transfer from single-image models via RL infeasible. To overcome this limitation and effectively jointly optimize planning and reflection, we propose the RPCO strategy.

\section{RPCO Methodology}
\label{sec:rpcotraining}
In this section, we introduce the RPCO strategy, designed to achieve the synergistic optimization of planning and reflection capabilities, which is a prerequisite for robust performance across both single-image and multi-image tasks.
To address the optimization asymmetry discussed in Theorem 3.1, RPCO adopts a 'decouple-then-fuse' paradigm. We first require a stable initialization via SFT to establish foundational behaviors, followed by RL to enforce long-horizon consistency and co-adaptation.
The efficacy of this initialization hinges on the quality of the SFT data. As demonstrated in Tab.~\ref{tab:single_image}, the Strong-Reflection model exhibits superior reflection capabilities compared to Gemini2.5Pro, while the latter excels in complex global planning. Consequently, we construct our training data by distilling high-quality reflection trajectories from the Strong-Reflection model and robust planning trajectories from Gemini2.5Pro.
In summary, as illustrated in Fig.~\ref{fig:framework} (b), the RPCO strategy is instantiated as a progressive three-stage training framework: (i) We first derive a Strong-Reflection model via SFT followed by single-image RL, explicitly honing visual diagnostic capabilities. (ii) We synthesize the VCR-SFT dataset by amalgamating distilled reflection trajectories (from Stage 1) with expert planning traces (from Gemini2.5Pro); subsequent fine-tuning on this composite corpus establishes a robust foundation with balanced priors. (iii) Finally, we leverage the VCR-RL dataset to conduct multi-task RL within multi-image workflows, facilitating the synergistic co-optimization of planning and reflection.

\subsection{Reflection on Single-Image Tasks}
We initially gather UTPCR trajectories from Gemini2.5Pro for SFT, establishing foundational reflection behaviors. Subsequently, we apply GRPO guided by the visual reflection reward (Eq.~\ref{reflect_reward}) and rule-based rewards to refine output fidelity.
\paragraph{Reflection Reward $R_{\text{reflect}}$.}
For reflection, we decompose each user request into $K$ evaluation checkpoints $\{q_1,\dots,q_K\}$ (e.g., subject correctness, style consistency, attribute accuracy, scene matching, action depiction, text accuracy) and let $\mathcal{I}$ denote the final accepted image set after reflection. A VLM-based judge (Qwen3VL32B) evaluates whether $\mathcal{I}$ satisfies each checkpoint $q_k$, yielding binary decisions $c_k \in \{\text{accept},\text{refuse}\}$. The Reflection Reward is defined as the satisfied-checkpoint ratio:
\begin{equation}
R_{\text{reflect}} = \frac{1}{K} \sum_{k=1}^{K} \mathbb{1}_{(c_k = \text{accept})}.
\label{reflect_reward}
\end{equation}
\vspace{-4mm}
\paragraph{Rule-based Rewards.}
We additionally use:
(i) \textbf{Format Reward} $R_{\text{format}}$: checks UTPCR tag presence, uniqueness, order, and non-emptiness across turns, penalizing trajectories with malformed structures.
(ii) \textbf{Tool Call Reward} $R_{\text{tool}}$: assigns a discrete score $\in\{0,0.1,0.8,1\}$ based on whether tool calls succeed, whether failures are self-corrected, and whether at least some calls succeed.
(iii) \textbf{Result Reward} $R_{\text{result}}$: binary success based on whether the number of generated images matches the requested counts.

\paragraph{Total Reward $R_{\text{total}}$.}
For a trajectory, the overall reward is
\begin{equation}
R_{\text{total}} = \frac{1}{|\mathcal{W}|} \sum_{i \in \mathcal{W}} w_i \cdot R_i,
\label{totalreward}
\end{equation}
where $\mathcal{W} = \{\text{reflection},  \text{format}, \text{tool}, \text{result}\}$ is the set of reward dimensions.
$w_i$ is the weight of dimension $i$ (set to 1 by default),
Each $R_i \in [0,1]$ is a normalized sub-reward.
{Reflection Reward} $R_{\text{reflect}}$ focuses on reflection visual quality.
{Format / Tool / Result Rewards} $R_{\text{format}}, R_{\text{tool}}, R_{\text{result}}$ are lightweight, rule-based components ensuring UTPCR structural correctness, successful tool execution, and basic output constraints, respectively.

Through this training phase, we obtain the Strong-Reflection model, which exhibits reflection capabilities superior to Gemini2.5Pro, thereby laying a solid foundation for acquiring high-quality reflection data.

\subsection{Advantage-complementary SFT}
Then, we adopt an advantage-complementary SFT strategy to achieve a high-quality performance initialization.
First, we construct an SFT corpus that combines:
(i) \textit{Reflection-strong single-image trajectories} generated by the Strong-Reflection model, which provide high-quality reflection patterns with clear error diagnosis and correction;
(ii) \textit{Planning-strong multi-image trajectories} generated by Gemini2.5Pro, which exhibit coherent global plans and high plan scores.
Then, these trajectories form the UTPCR-style dataset used for SFT, resulting in the \textit{Reflection-Plan SFT} model.
As shown in Tab.~\ref{tab:multi_image}, this model achieves a substantial improvement in multi-image planning (plan score 0.9452) while restoring balanced reflection behavior.

\subsection{Multi-task RL \& Co-Optimization}
Building upon the Reflection-Plan SFT model, which is initialized with balanced planning and reflection capabilities, we further perform multi-task RL on both single-image and multi-image tasks to achieve the co-optimization of high-level reasoning and visual self-correction. Upon completion of this training phase, we obtain the VisionCreator-R1 model.
The multi-dimensional reward design for GRPO is as follows:
\paragraph{Plan Reward $R_{\text{plan}}$.}
Given a user requirement $Q$ and an agent-generated plan $P$, we employ an external LLM-based evaluator $f_{\text{eval}}(Q,P) \in \{0,1,\dots,N\}$ to assess (i) requirement completeness, (ii) logical coherence of the sub-task sequence, and (iii) tool–goal matching. The normalized Plan Reward is:
\begin{equation}
R_{\text{plan}} = \frac{f_{\text{eval}}(Q,P)}{N}.
\end{equation}
This reward is applied at the plan stage and is the primary signal driving the improvement of long-horizon, reference-aware planning in multi-image settings.

\paragraph{Total Reward $R_{\text{total}}$.}
We adopt the unified reward formulation consistent with Eq.~\ref{totalreward}, but expand the component set to $\mathcal{W} = \{\text{reflection}, \text{plan}, \text{format}, \text{tool}, \text{result}\}$ to facilitate the synergistic co-optimization of planning and reflection capabilities. In this setup, the Plan Reward $R_{\text{plan}}$ explicitly evaluates the rationality of the generated plan. In practice, Plan Reward $R_{\text{plan}}$ and Reflection Reward $R_{\text{reflect}}$ serve as the dominant learning signals driving high-level reasoning and visual self-correction, respectively, while the remaining terms function primarily as structural regularizers to ensure stability during long-horizon training.

\begin{figure}[ht]
\vspace{-2mm}
\centering
\includegraphics[width=0.67\linewidth]{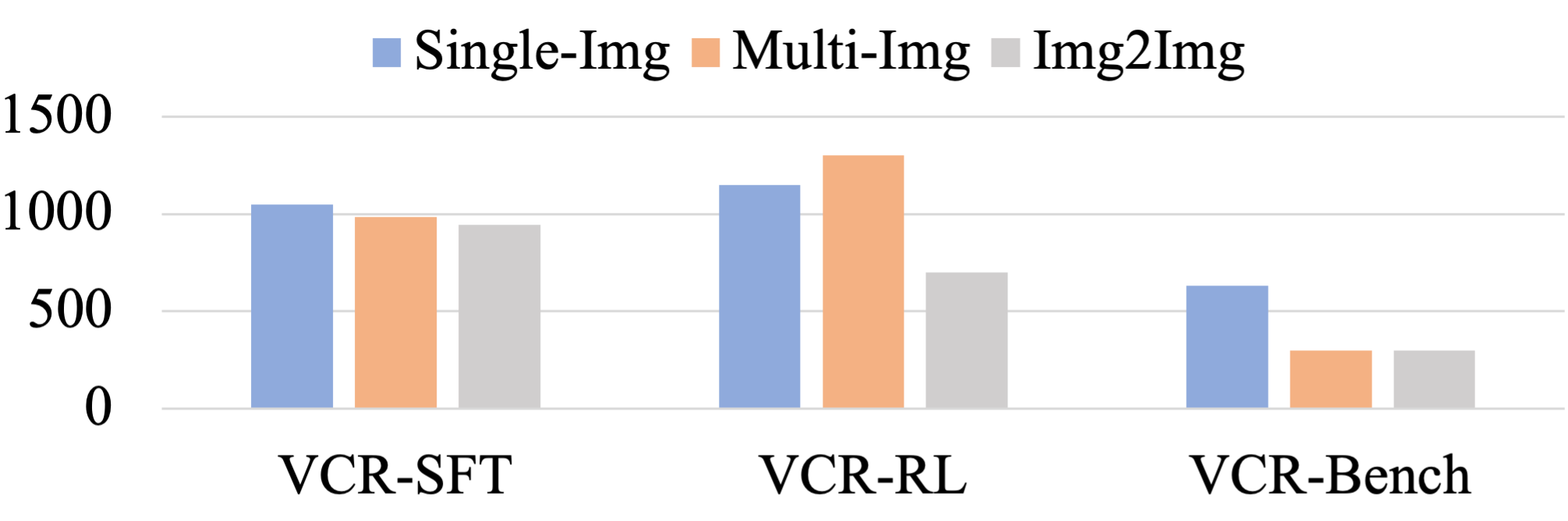}
\vspace{-2mm}
\caption{Task distribution of the VCR dataset.}
\vspace{-4mm}
\label{fig:dataset_task}
\end{figure}

\section{VCR Dataset}
\label{sec:vcr_dataset}
We construct VCR-SFT and VCR-RL datasets to support RPCO methodology. And we propose VCR-Bench for standardized assessment.
Fig.~\ref{fig:dataset_task} shows the task distribution of the VCR dataset, which includes the tasks of Single-Img, Multi-Img, and Img2Img.

\begin{figure}[ht]
\vspace{-2mm}
\centering
\includegraphics[width=0.75\linewidth]{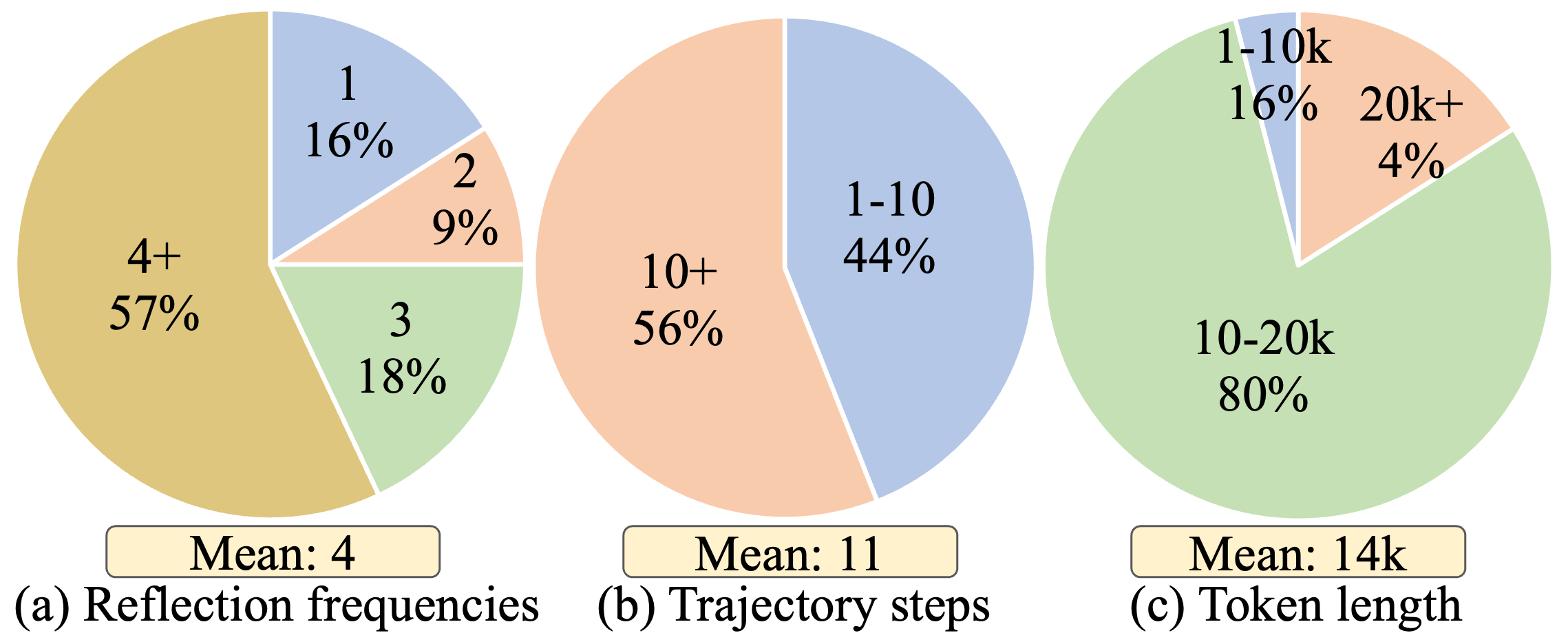}
\vspace{-1mm}
\caption{VCR-SFT dataset distribution.}
\vspace{-2mm}
\label{fig:dataset}
\end{figure}

\subsection{VCR-Data: SFT and RL Resources}
\label{sec:vcr_data}
\textbf{VCR-SFT.}
VCR-SFT is designed for supervised fine-tuning and provides explicit UTPCR-format trajectories, with explicit tags for understanding, thinking, planning, creation, and reflection.
VCR-SFT dataset distribution is presented in Fig.~\ref{fig:dataset}.
To ensure the quality of the SFT data, we strictly filter the trajectories by retaining only those that achieve perfect scores with \{$R_{\text{plan}} = 1$, $R_{\text{format}} = 1$, $R_{\text{tool}} = 1$\}.
More importantly, VCR-SFT provides effective supervision for improving reflection capability and planning capability.
Specifically, VCR-SFT consists of two complementary subsets:
(i) \textit{Single-image reflection-strong trajectories}, collected from our Strong-Reflection model.
They provide high-quality supervision for grounded visual reflection behaviors.
(ii) \textit{Multi-image planning-strong trajectories}, collected from Gemini2.5Pro on multi-image tasks.
These trajectories emphasize strong global planning.

\textbf{VCR-RL.}
VCR-RL is designed for RL.
Each VCR-RL sample contains a user query and a list of reflection evaluation checkpoints that specify desired properties of the \textit{final visual outputs}.
During RL training, the \textit{reflection reward} is computed by applying a VLM-based judge to the post-reflection images against the query-specific reflection checkpoints, i.e., each query's checkpoints are tailored to its specific requirements.
This diversity trains the agent to adapt its reflection and correction strategies to varied task demands, enhancing robustness and generalization.

\begin{wraptable}{r}{0.55\columnwidth}
\vspace{-5mm}
\centering
\renewcommand{\tabcolsep}{2.2pt}
\renewcommand{\arraystretch}{1.1}
\small
\caption{Comparison of semantic consistency (G\_SC), perceptual quality (G\_PQ), overall score (G\_O) on GEdit-Bench.}
\vspace{-2mm}
\begin{tabular}{l|ccc}
\hline
\bf Model& G\_SC & G\_PQ & G\_O \\
\hline
Gemini 2.0~\citep{googleGemini2} & 6.73 & 6.61 & 6.32 \\
BAGEL~\citep{deng2025bagel} & 7.36 & 6.83 & 6.52  \\
FLUX.1 Kontext Pro~\citep{labs2025kontext} & 7.02 & \textbf{7.60} & 6.56  \\
Step1X-Edit~\citep{liu2025step1x} & \textbf{7.66} & 7.35 & 6.97  \\
\hline
Qwen-Image-Fast &7.26&7.34& 7.03\\
VisionCreator-R1 &{7.60}&7.33&\textbf{7.23}\\
\hline
\end{tabular}
\label{tab:gedit}
\vspace{-9mm}
\end{wraptable}

\subsection{VCR-Bench: Evaluation Suite}
\label{sec:vcr_bench}
VCR-Bench reports final task performance, also supports trajectory-level analysis by computing reflection and plan statistics from inference traces.
Each benchmark query contains a user instruction and a set of evaluation checkpoints, and we employ \textit{Gemini2.5Pro} as a VLM-based judge for automatic evaluation.

\section{Experiment}
\label{sec:exp}

\subsection{Baselines and Implementations}
The used image generation and editing tool Qwen-Image-Fast is Qwen-Image-Edit-2509-Lightning-4steps-V1.0 \cite{qwenimagelightning2025}.
All compared agent models (i.e., Qwen3VL32B, Gemini2.5Pro) integrate the same UTPCR-format system prompt as shown in Appendix.
Our VsionCreator-R1 adopts Qwen3VL32B as the base model, and sets $N=6$ for plan reward.

\label{sec:overall_results}
\begin{wraptable}{r}{0.55\columnwidth}
\vspace{-2mm}
\centering
\renewcommand{\tabcolsep}{2.1pt}
\renewcommand{\arraystretch}{1.1}
\small
\caption{Performance Comparison on VCR-Bench.}
\label{tab:sota}
\begin{tabular}{l|c|c|c}
\hline
\textbf{Model} & \textbf{Single-Img} & \textbf{Multi-Img} & \textbf{Img2mg} \\
\hline
\multicolumn{4}{l}{\textit{Evaluated by Gemini2.5Pro}}\\
\hline
Qwen-Image-Fast & 0.505 & 0.480 & 0.672 \\
Qwen3VL32B & 0.500 & 0.601 & 0.806 \\
Gemini2.5Pro & 0.515 & 0.649 & 0.816 \\
VisionCreator-R1 & 0.532 & 0.700 & 0.836 \\
\hline
\multicolumn{4}{l}{\textit{Evaluated by Human}}\\
\hline
VC-R1 vs. Gemini & +14.8\% & +9.3\% & +5.8\% \\
\hline
\end{tabular}
\vspace{-2mm}
\end{wraptable}

\subsection{Results on GEdit-Bench}
Tab.~\ref{tab:gedit} reports the results on the \textbf{GEdit-Bench}~\citep{liu2025step1x}, which evaluates image editing models on user instructions across 11 diverse categories. The results show that our agent, VisionCreator-R1, achieves the highest overall score (7.23). Notably, VisionCreator-R1 significantly improves the semantic consistency score compared to the base tool (7.60 vs. 7.26), demonstrating that our agent's reflection and planning abilities enable more accurate and intent-aligned editing.

\subsection{Results on VCR-Bench}
Tab.~\ref{tab:sota} summarizes the performance comparison on \textbf{VCR-Bench} across three representative task categories: single-image generation, multi-image generation, and image-to-image editing.

\noindent \textbf{Automatic evaluation by Gemini2.5Pro.}
For automatic evaluation, we employ \textit{Gemini2.5Pro} as a strong VLM-based judge to score each model output against multiple predefined evaluation checkpoints.
Under this protocol, \textit{VisionCreator-R1} consistently achieves the best performance across all three task categories.
Notably, the gain on \textit{multi-image tasks} is the most pronounced, where VisionCreator-R1 outperforms Gemini2.5Pro itself by a large margin (0.700 vs.\ 0.649).
This result indicates that explicit reflection and reflection--plan co-optimization enable more reliable long-horizon reasoning and error correction than purely plan-driven generation.

\begin{table*}
\centering
\renewcommand{\tabcolsep}{2.8pt}
\renewcommand{\arraystretch}{1.1}
\small
\caption{Reflection-Plan experiment on VCR-Bench. Reflection Quality and Plan Score are calculated on Multi-Img task traces.}
\label{tab:exp}
\vspace{-2mm}
\begin{tabular}{l|c|c|c|c}
\hline
\textbf{Model} & \textbf{Single-Img} & \textbf{Multi-Img} & \textbf{Reflection Quality (U/G/O)} & \textbf{Plan Score} \\
\hline
Qwen-Image-Fast & 0.505 & 0.480 & -- & -- \\
\hline
Gemini2.5Pro & 0.515 & 0.649 & 61.6\%/31.8\%/6.6\% & 0.9667 \\
Qwen3VL32B & 0.500 & 0.601 & 75.4\%/21.4\%/3.3\% & 0.8125 \\
\hline
Strong-Reflection & \textbf{0.543} & 0.565 & 68.6\%/21.6\%/9.8\% & 0.8285 \\
Reflection-Plan Conflict & 0.539 & 0.617 & 73.9\%/16.5\%/9.7\% & 0.8431 \\
\hline
Reflection-Plan SFT & 0.523 & 0.647 & 64.0\%/\textbf{32.4\%}/3.6\% & 0.9452 \\
VisionCreator-R1 & 0.532 & \textbf{0.700} & 63.5\%/31.0\%/5.5\% & \textbf{0.9746} \\
\hline
\end{tabular}
\vspace{-2mm}
\end{table*}

\begin{wrapfigure}{r}{0.5\textwidth}
\vspace{-2mm}
\centering
\includegraphics[width=0.9\linewidth]{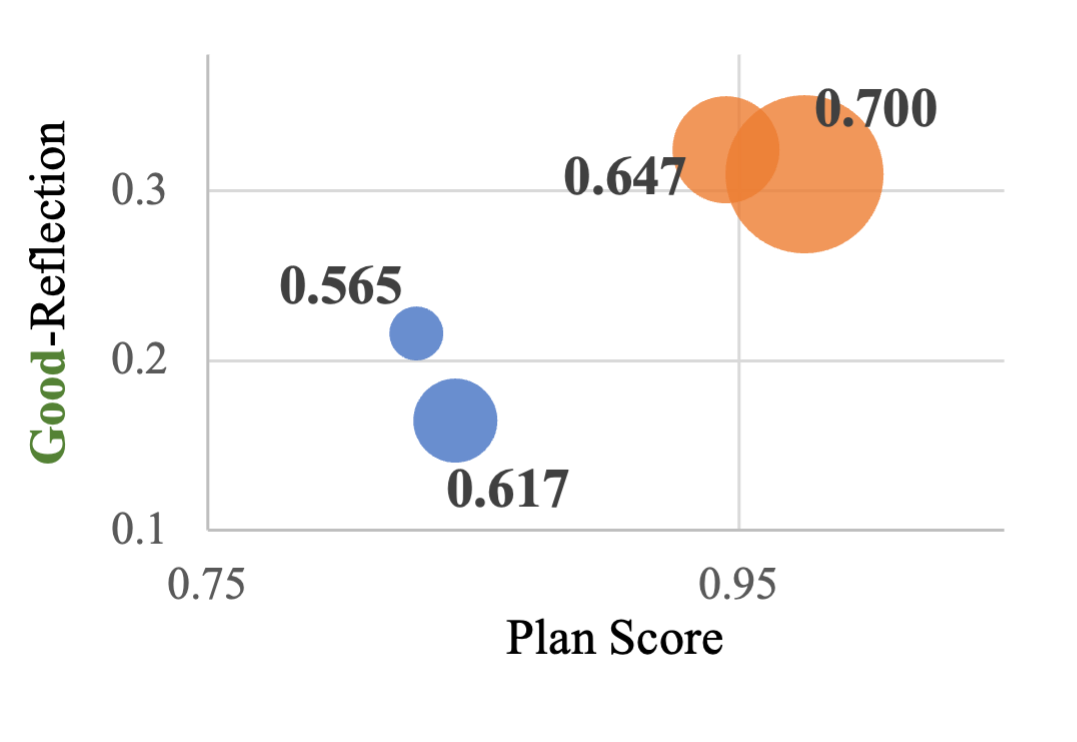}
\vspace{-5mm}
\caption{Plan-Reflection with Performance.}
\vspace{-2mm}
\label{fig:exp}
\end{wrapfigure}

\noindent \textbf{Human evaluation.}
To validate that automatic scores align with human preference, we conduct a pairwise human evaluation between \textit{VisionCreator-R1} and \textit{Gemini2.5Pro}.
Human annotators are asked to judge which output better satisfies the user intent across all evaluation checkpoints.
As shown in Tab.~\ref{tab:sota}, VisionCreator-R1 is preferred over Gemini2.5Pro in \textbf{14.8\%} of single-image task, \textbf{9.3\%} of multi-image task, and \textbf{5.8\%} of image-to-image task.
These results confirm that the improvements measured by VLM-based evaluation correspond to tangible quality gains perceived by human users.
The consistency between automatic and human evaluations supports the reliability of VCR-Bench as a testbed for evaluating visual generation.

\subsection{Ablation Studies on VCR-Bench}

\noindent \textbf{Evaluation protocol and statistics in Tab.~\ref{tab:exp}.}
For \textbf{single-image tasks}, we report only the final task performance, since earlier experiments have already established a strong positive correlation between reflection quality and final task performance.
In contrast, \textbf{multi-image tasks} involve long-horizon decision making with multiple intermediate steps.
Therefore, in addition to final task performance, we report \textit{trajectory-level statistics} computed from multi-image inference traces, including reflection behavior analysis and plan scores.
All reflection and planning statistics in Tab.~\ref{tab:exp} are computed exclusively from multi-image task traces.

\noindent \textbf{Discussion.}
In Tab.~\ref{tab:exp}, from \textit{Strong-Reflection model} to \textit{Reflection-Plan Conflict model}, we observe that when planning capability is weak, directly optimizing reflection via RL leads to reflection degradation.
According to Theorem~\ref{thm:variance_asymmetry}, weak planning produces more invalid intermediate images, triggering frequent reflection and additional tool invocations.
This lengthens trajectories and amplifies diffusion-induced stochasticity, causing the trajectory variance $\Sigma_\tau$ to dominate the gradient estimator.
Under this high-variance regime, GRPO cannot reliably attribute rewards to reflection actions, resulting in unstable optimization.
This explains why strong planning prior to reflection-optimized RL is necessary.
Consistent with this analysis, initializing from \textit{Reflection-Plan SFT} substantially stabilizes training, leading to the strongest overall performance in \textit{VisionCreator-R1}.
Fig.~\ref{fig:exp} demonstrates that improving reflection and planning capabilities achieves obvious performance gains in multi-image tasks.

\begin{figure*}[ht]
\vspace{-2mm}
\centering
\includegraphics[width=0.99\textwidth]{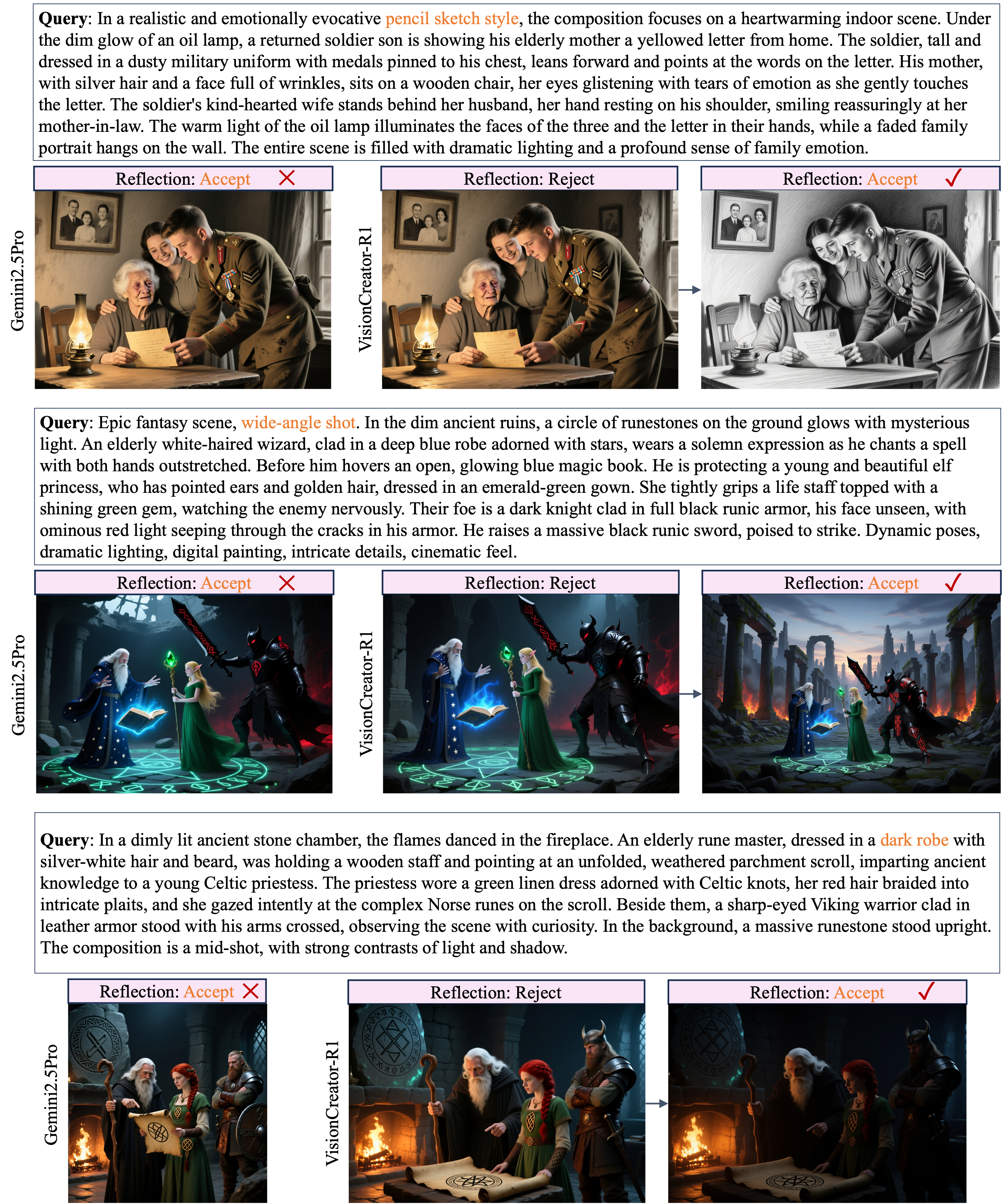}
\vspace{-1mm}
\caption{Visualization comparisons.}
\vspace{-2mm}
\label{fig:visual_single}
\end{figure*}

\section{Conclusion}
We study reflection as a trainable capability in native visual generation agents and show that naïvely optimizing reflection via RL is fundamentally limited by a structural variance asymmetry.
By isolating reflection in low-noise single-image settings, we demonstrate that reflection is learnable and substantially improves visual quality.
We further show that in multi-image tasks, planning and reflection exhibit asymmetric optimization regimes, motivating a decoupled-then-fused strategy.
Guided by this insight, we propose RPCO and introduce VisionCreator-R1, which consistently outperforms Gemini2.5Pro across single-image, multi-image, and image-to-image benchmarks.
Our results establish principled guidelines for training reflection-aware visual agents under long-horizon stochastic environments.
We further release {VCR-SFT}, {VCR-RL}, and {VCR-Bench} to support future research on reflection-aware visual generation.

\clearpage
\setcounter{section}{0}
\renewcommand{\thesection}{\Alph{section}}
\renewcommand{\thesubsection}{\thesection.\arabic{subsection}}

\section{Rigorous Derivation of Gradient Variance in GRPO}

\subsection{The GRPO Gradient Estimator}

The gradient of the GRPO objective is estimated by sampling a group of outputs $\{o_i\}_{i=1}^G$. To analyze the optimization difficulty, we focus on the stochastic gradient estimator $\hat{g}$ for a single token generation step $t$ in a single trajectory $i$.

Let $s$ be the context $(q, o_{i, <t})$ and $a$ be the current token $o_{i,t}$. The estimator $\hat{g}$ is (assume $\pi_{\theta} =  \pi_{old}$ for simplified analysis):

\begin{equation}
    \hat{g} = \left[ \hat{A}_{i,t} + \beta \left( \frac{\pi_{ref}(a|s)}{\pi_\theta(a|s)} - 1 \right) \right] \nabla_\theta \log \pi_\theta(a|s),
\end{equation}
where $\hat{A}_{i,t}$ denotes the advantage derived from the group rewards ($\hat{A}_{i,t}=\frac{r_i - \mu_r}{\sigma_r}$), in which $r_i$ crucially depends on the entire final trajectory $\tau$. The term involving $\beta$ represents the KL regularization; for a fixed policy state $\theta$ and reference policy $\pi_{ref}$, given action $a$, this term is deterministic. We denote this deterministic scalar part as $D(a,s) = \beta \left(\frac{\pi_{ref}}{\pi_\theta} - 1\right)$. 
Consequently, the stochasticity inherent in $\hat{g}$ stems from two sources: the policy-driven action sampling ($a \sim \pi_\theta$) for token selection, and the trajectory stochasticity ($\tau|a$). The latter encapsulates the variability of the image diffusion process, which governs the advantage estimate $\hat{A}_{i,t}$.

\subsection{Variance Decomposition via Law of Total Variance}

We decompose the total variance $\Sigma := \text{Var}(\hat{g})$ by recursively applying the Law of Total Variance.

First, we decompose the variance with respect to the state $s$:
\begin{equation}
\begin{aligned}
\Sigma = \text{Var}_{s,a,\tau}(\hat{g}) = \mathbb{E}_s \left[ \text{Var}_{a,\tau|s}(\hat{g}) \right] + \text{Var}_s \left( \mathbb{E}_{a,\tau|s}[\hat{g}] \right).
\end{aligned}
\end{equation}

Next, we further decompose the inner term $\text{Var}_{a,\tau|s}(\hat{g})$ by conditioning on the action $a$, given the state $s$:
\begin{equation}
\begin{aligned}
\Sigma &= \mathbb{E}_s \left[ \mathbb{E}_{a|s} \left[ \text{Var}_{\tau|s,a}(\hat{g}) \right] + \text{Var}_{a|s} \left( \mathbb{E}_{\tau|s,a}[\hat{g}] \right) \right] + \underbrace{\text{Var}_s \left( \mathbb{E}_{a,\tau|s}[\hat{g}] \right)}_{\Sigma_s} \\
&= \underbrace{\mathbb{E}_{s,a} \left[ \text{Var}_{\tau|s,a}(\hat{g}) \right]}_{\Sigma_\tau} + \underbrace{\mathbb{E}_s \left[ \text{Var}_{a|s} \left( \mathbb{E}_{\tau|s,a}[\hat{g}] \right) \right]}_{\Sigma_a} + \Sigma_s,
\end{aligned}
\end{equation}
where $\Sigma_\tau$ represents the variance attributable to environmental dynamics and reward stochasticity, conditioned on a specific state and action. $\Sigma_a$ denotes the variance resulting from the policy's stochastic action sampling given a state; notably, this is the specific component minimized by the baseline in GRPO. Finally, $\Sigma_s$ captures the variability stemming from the underlying distribution of states (i.e., input prompts). 
Since $\Sigma_s$ measures the variance of the true gradients across different prompts and is independent of the algorithm design, we omit it from our consideration. Thus, $\Sigma = {\Sigma_\tau}+{\Sigma_a}$.

\subsection{Step-by-Step Derivation of Terms}
First, we analyze the variance of the gradient estimator $\hat{g}$ conditioned on a specific action $a$, denoted as $\Sigma_\tau$. Given a fixed action $a$, both the function $\mathbf{v} = \nabla_\theta \log \pi_\theta(a|s)$ and the KL-divergence term $D(a,s)$ are treated as constants. Consequently, the conditional variance arises solely from the Advantage term $\hat{A}_{i,t}$, which fluctuates due to the stochastic nature of the trajectory rewards $r_i$ (e.g., inherent noise in the diffusion process).

Substituting the expression for $\hat{g}$ into the conditional variance formulation yields:
\begin{equation}
\begin{aligned}
    \text{Var}_{\tau|a}(\hat{g}) &= \text{Var}_{\tau|a} \left( (\hat{A}_{i,t} + D(a,s)) \cdot \mathbf{v} \right) \\
    &= \mathbf{v}\mathbf{v}^\top \cdot \text{Var}_{\tau|a}(\hat{A}_{i,t}).
\end{aligned}
\end{equation}
The total trajectory variance $\Sigma_\tau$ is obtained by taking the expectation over the action space:
\begin{equation}
    \Sigma_\tau = \mathbb{E}_a \left[ \lVert \nabla_\theta \log \pi_\theta(a|s) {\rVert}^2 \cdot \text{Var}_{\tau|a}(\hat{A}_{i,t}) \right].
\end{equation}

This derivation reveals a critical challenge in image-generating tasks, particularly when employing reflection mechanism. Since the advantage $\hat{A}_{i,t}$ depends on the stochastic outcomes of the diffusion model, the term $\text{Var}_{\tau|a}(\hat{A}_{i,t})$ can become exceedingly large. Even when the agent selects a consistent reflection token $a$, the varying quality of the generated images introduces significant noise into the reward signal, thereby amplifying the overall variance of the estimator. Moreover, the synthesis of multiple images introduces significantly higher environmental dynamics and compounded stochasticity. This amplified variance renders the optimization landscape far more volatile, making the training process significantly more difficult than in standard single-image tasks.

Next, we analyze the expectation term $\mathbb{E}_{\tau|a}[\hat{g}]$.
\begin{equation}
\begin{aligned}
    \mathbb{E}_{\tau|a}[\hat{g}] &= \mathbb{E}_{\tau|a} \left[ (\hat{A}_{i,t} + D(a,s)) \cdot \nabla_\theta \log \pi_\theta(a|s) \right] \\
    &= \left( \mathbb{E}_{\tau|a}[\hat{A}_{i,t}] + D(a,s) \right) \cdot \nabla_\theta \log \pi_\theta(a|s).
\end{aligned}
\end{equation}
Let $\mathcal{A}(s,a) = \mathbb{E}_{\tau|a}[\hat{A}_{i,t}]$ be the {expected advantage} of taking action $a$. This represents the quality of the action, averaged over environmental noise.

Thus, $\Sigma_a$ is the variance of this expected gradient across different actions:
\begin{equation}
    \Sigma_a = \text{Var}_a \left( \left[ \mathcal{A}(s,a) + \beta \left( \frac{\pi_{ref}(a|s)}{\pi_\theta(a|s)} - 1 \right)  \right] \cdot \nabla_\theta \log \pi_\theta(a|s) \right).
\end{equation}
This term encapsulates the informative learning signal, quantifying the gradient variability induced by the agent's exploration across the action space $a$.

Finally, combining the terms:

\begin{equation}
    \text{Var}(\hat{g}) = \underbrace{\mathbb{E}_a [ \lVert\nabla \log \pi{\rVert}^2 \cdot \text{Var}_{\tau|a}(\hat{A}_{i,t}) ]}_{\Sigma_\tau} + \underbrace{\text{Var}_a \left( \left[ \mathcal{A}(s,a) + \beta \left( \frac{\pi_{ref}(a|s)}{\pi_\theta(a|s)} - 1 \right)  \right] \cdot \nabla_\theta \log \pi_\theta(a|s) \right)}_{\Sigma_a}
\end{equation}

We identify a critical disparity in the variance composition between the planning and reflection objectives, which we term \textit{structural asymmetry}.

For the {planning} mechanism ($R_{plan}$), the reward evaluation is typically deterministic. Conditioned on a fixed action $a$, the reward signal remains invariant, rendering the conditional variance negligible:
\begin{equation}
    \text{Var}_{\tau|a}(\hat{A}_{i,t}) \approx 0 \quad \implies \quad \Sigma_\tau \approx 0.
\end{equation}
Under these conditions, the total variance is dominated by the variance term $\Sigma_a$. Consequently, the optimization process remains stable, as the gradient primarily reflects the useful learning signal derived from action exploration.

In contrast, the {reflection} mechanism ($R_{reflect}$) is inherently contingent upon the stochastic outcomes of the diffusion model. This compounded stochasticity causes the conditional variance to become excessively large:
\begin{equation}
    \text{Var}_{\tau|a}(\hat{A}_{i,t}) \gg 0 \quad \implies \quad \Sigma_\tau \gg \Sigma_a.
\end{equation}
Consequently, the gradient estimator becomes dominated by the trajectory noise $\Sigma_\tau$ rather than the informative signal $\Sigma_a$. This imbalance precipitates a collapse in the Signal-to-Noise Ratio, rendering the optimization landscape highly volatile. Under such conditions, the stable acquisition of gradients is obstructed, making the reflection capabilities difficult to optimize.

\section{VisionCreator-R1 Reward Design}
\label{sec:reward_design}

In this section, we describe the multi-dimensional reward system used to train VisionCreator-R1 with reinforcement learning. The reward mechanism decomposes the supervision signal into complementary components that jointly guide the agent through high-level planning, structural compliance, tool execution, result satisfaction, and visual reflection.

\subsection{Overall Reward Function}
\label{subsec:overall_reward}

We aggregate multiple sub-rewards via a weighted average to ensure the agent receives learning signals at all stages of the UTPCR pipeline. The total reward is calculated as
\begin{equation}
R_{\text{total}} = \frac{1}{|\mathcal{W}|} \sum_{i \in \mathcal{W}} w_i \cdot R_i,
\end{equation}
where $\mathcal{W}$ represents the set of reward dimensions comprising plan, format, tool, result, and reflection components. The weight $w_i$ for each dimension is set to 1 by default, while each sub-reward $R_i$ is normalized to the range $[0,1]$. Specifically, the Plan Reward $R_{\text{plan}}$, Format Reward $R_{\text{format}}$, and Reflection Reward $R_{\text{reflect}}$ take values within $[0, 1]$. The Tool Call Reward $R_{\text{tool}}$ assumes discrete values from the set $\{0, 0.1, 0.8, 1\}$, and the Result Reward $R_{\text{result}}$ is a binary value in $\{0, 1\}$. This design balances the training process, preventing the agent from being driven solely by the final output quality.

\subsection{Plan Reward}
\label{subsec:plan_reward}

The Plan Reward evaluates the quality of the agent's task plan, ensuring it correctly understands user requirements and proposes a coherent, tool-aware execution strategy. Let the user requirement be $Q$ and the agent's plan be $P$. We utilize an external LLM-based evaluator $f_{\text{eval}}(Q, P)$ that returns a discrete score in the range $\{1, 2, \dots, N\}$ with $N=6$. This score is normalized to obtain the final reward
\begin{equation}
R_{\text{plan}} = \frac{f_{\text{eval}}(Q, P)}{N}.
\end{equation}
The evaluator $f_{\text{eval}}$ jointly considers three critical aspects: requirement completeness, which checks if the plan covers the correct output types and counts; logical coherence, which assesses the reasonableness of sub-task decomposition and ordering; and tool matching, which verifies that the chosen tools are appropriate for the intended sub-goals.

\subsection{Format Reward}
\label{subsec:format_reward}

The Format Reward assesses whether the agent’s outputs strictly follow the predefined UTPCR tag schema, a crucial factor for stable training and reliable tool orchestration. We enforce three canonical patterns depending on the dialogue turn. The first turn must follow the sequence of thinking, planning, thinking, tool calling, and tool result processing. Middle turns consist of reflection, thinking, tool calling, and tool result processing. The final turn requires a sequence of reflection, thinking, and the final answer.

For a trajectory with $T$ turns, let $s_t \in [0,1]$ denote the format score of turn $t$. The trajectory-level reward is defined as the minimum score across all turns, $R_{\text{format}} = \min_{t} s_t$, ensuring that a single badly formatted turn penalizes the entire interaction. For each individual turn, the score is calculated as
\begin{equation}
s_t = \frac{N_{\text{valid}} + \mathbb{1}_{\text{order}}}{N_{\text{required}} + 1},
\end{equation}
where $N_{\text{valid}}$ is the number of tags passing validation, $N_{\text{required}}$ is the expected number of tags for that turn type, and $\mathbb{1}_{\text{order}}$ is an indicator function that equals 1 if the tag order is fully correct. A tag is considered valid if and only if it exists, appears exactly once, and contains non-empty content.

\subsection{Tool Call Reward}
\label{subsec:tool_reward}

The Tool Call Reward measures the correctness and robustness of tool usage based on execution success. It encourages consistent success while providing partial credit for self-correcting behavior. We define a success indicator $success(r)$ which equals 1 if the tool execution result $r$ indicates success, and 0 otherwise. For a trajectory with $n$ tool calls and results $(r_1, \dots, r_n)$, we define three conditions: $\alpha = \prod_{i=1}^{n-1} success(r_i)$ represents success in all intermediate calls; $\beta = success(r_n)$ represents success in the final call; and $\gamma = \max_{i} success(r_i)$ indicates at least one successful call occurred.

The Tool Call Reward is assigned according to a piecewise schedule:
\begin{equation}
R_{\text{tool}} =
\begin{cases}
1.0, & \text{if } \alpha = 1 \land \beta = 1, \\
0.8, & \text{if } \alpha = 0 \land \beta = 1, \\
0.1, & \text{if } \beta = 0 \land \gamma = 1, \\
0.0, & \text{otherwise}.
\end{cases}
\end{equation}
This scheme assigns the maximum reward of 1.0 for consistently correct usage. It grants 0.8 for self-corrected paths where early failures are followed by a successful final call, 0.1 for partial progress where failures persist but some attempts succeeded, and 0.0 for complete failure.

\subsection{Result Reward}
\label{subsec:result_reward}

The Result Reward validates whether the final task output satisfies basic quantitative constraints, specifically the requested numbers of images and videos. Let $N_{\text{img}}^{\text{exp}}$ and $N_{\text{img}}^{\text{act}}$ be the expected and actual image counts, and $N_{\text{vid}}^{\text{exp}}$ and $N_{\text{vid}}^{\text{act}}$ be the expected and actual video counts. The reward is defined as
\begin{equation}
R_{\text{result}} =
\mathbb{1}_{(N_{\text{img}}^{\text{act}} = N_{\text{img}}^{\text{exp}})}
\cdot
\mathbb{1}_{(N_{\text{vid}}^{\text{act}} = N_{\text{vid}}^{\text{exp}})}.
\end{equation}
This formula yields a value of 1 only if both image and video counts strictly match the requirements, and 0 otherwise. This binary signal enforces minimum task completion criteria independent of finer-grained visual quality.

\subsection{Reflection Reward}
\label{subsec:reflection_reward}

The Reflection Reward leverages a VLM to evaluate whether the \textit{post-reflection} images satisfy a set of pre-defined evaluation points derived from the user requirement. We decompose the user requirement into $K$ checkpoints $\{q_1, \dots, q_K\}$. Given the final image set $\mathcal{I}$, a VLM-based judge $g(q_k, \mathcal{I})$ determines whether each checkpoint $q_k$ is accepted or refused. Letting $c_k$ denote the judgment result, the reward is calculated as the fraction of satisfied checkpoints:
\begin{equation}
R_{\text{reflect}} = \frac{1}{K} \sum_{k=1}^{K} \mathbb{1}_{(c_k = \text{accept})}.
\end{equation}
The evaluation points typically cover subject correctness, style consistency, attribute accuracy, scene matching, action depiction, and text accuracy.

\subsection{Reward Properties and Interactions}
\label{subsec:reward_properties}

The proposed reward system integrates a mixture of discrete and continuous signals to stabilize RL training while providing rich supervision over structure and content. $R_{\text{format}}$ serves as a continuous signal, whereas $R_{\text{plan}}$, $R_{\text{tool}}$, $R_{\text{result}}$, and $R_{\text{reflect}}$ provide discrete feedback steps. There are inherent dependencies among these dimensions: high-quality plans increase the likelihood of meeting basic output constraints, while correct tool usage is a prerequisite for achieving desired results. Furthermore, effective reflection combined with strong initial results leads to high post-reflection visual quality. Collectively, this multi-dimensional design covers the full lifecycle of a VisionCreator-R1 trajectory, enabling the Reflection--Plan Co-Optimization framework to balance improvements in planning and reflection capabilities.

\clearpage

\bibliography{colm2024_conference}
\bibliographystyle{colm2024_conference}
\end{document}